# پیما: پیکره برچسب خورده موجودیت‌های اسمی زبان فارسی

مهساسادات شهشهانی[1]     مهدی محسنی[1]     آزاده شاکری[1,*]     هشام فیلی[1]

1. دانشکده مهندسی برق و کامپیوتر، پردیس دانشکده‌های فنی دانشگاه تهران، تهران، ایران

چکیده:

در مسئله تشخیص موجودیت‌های اسمی، هدف طبقه‌بندی اسامی خاص متن با برچسب‌هایی همچون شخص، مکان، و سازمان است. این مسئله به عنوان یکی از گام‌های پیش‌پردازشی بسیاری از مسائل پردازش زبان طبیعی مطرح است. اگر چه در زبان انگلیسی پژوهش‌های زیادی در این حوزه انجام شده و سیستم‌ها به دقت بالای 90 درصد دست یافته‌اند، در زبان فارسی به دلیل نبود یک مجموعه داده‌ی استاندارد، پژوهش کمی در این زمینه انجام گرفته است. در این پژوهش به ساخت چنین مجموعه داده‌ای می‌پردازیم و آن را به صورت آزاد در اختیار پژوهشگران آینده قرار می‌دهیم. سپس با استفاده از این مجموعه داده به طراحی سیستم ترکیبی (آماری و قاعده‌محور) با استفاده از مدل CRF و نیز سیستمی مبتنی بر شبکه‌های عصبی یادگیری عمیق از نوع LSTM می‌پردازیم. نتایج آزمایش‌ها نشان می‌دهد که با استفاده از سیستم ترکیبی (آماری و لیست‌محور) به F1 برابر با 84درصد برای هفت برچسب شخص، مکان، سازمان، تاریخ، زمان، درصد و واحد پول می‌رسیم.

واژه‌های کلیدی: پیکره‌ی موجودیت‌های اسمی، روش قاعده‌محور، روش مبتنی بر یادگیری عمیق، روش میدان‌های شرطی تصادفی

\* نویسنده مسئول

# PEYMA: A Tagged Corpus for Persian Named Entities

Mahsa Sadat Shahshahani[1]  Mahdi Mohseni[1]     Azadeh Shakery[1,*]    Heshaam Faili[1]

1. School of Electrical and Computer Engineering, Faculty of Engineering, Tehran, Iran.

**Abstract:**
The goal in the named entity recognition task is to classify proper nouns of a text into classes such as person, location, and organization. This is an important preprocessing step in many natural language processing tasks such as question-answering and summarization. Although many research studies have been conducted in this area in English and the state-of-the-art NER systems have reached performances of higher than 90 percent in terms of F1 measure, there are very few research studies for this task in Persian. One of the main important causes of this may be the lack of a standard Persian NER dataset to train and test NER systems. In this research we create a standard, big-enough tagged Persian NER dataset which will be distributed for free for research purposes. In order to construct such a standard dataset, we studied standard NER datasets which are constructed for English researches and found out that almost all of these datasets are constructed using news texts. So we collected documents from ten news websites. Later, in order to provide annotators with some guidelines to tag these documents, after studying guidelines used for constructing CoNLL and MUC standard English datasets, we set our own guidelines considering the Persian linguistic rules. Using these guidelines, all words in documents can be labeled as person, location, organization, time, date, percent, currency, or other (words that are not in any other 7 classes). We use IOB encoding for annotating named entities in documents, like in most of the existing English standard NER datasets. Using this encoding, the first token of a named entity will label with B, next tokens of it (if exist) will label with I. Other words which are not part of any named entity, will label with O. Final corpus consists of 709 documents, which includes 302530 tokens. 41148 tokens out of these tokens are labeled as named entity and the others are labeled as O. In order to determine inter-annotator agreement, 160 documents were labeled by different annotators. Kappa statistic was estimated as 95% using words that are labeled as named entities. After creating the dataset, we use it to design a hybrid system for named entity recognition. Our hybrid system consists of a rule-based and a statistical part. The rule-based part consists of lists of some frequent named entities as well as some regular expressions based on Persian linguistic rules to identify named entities, and the statistical part is based on conditional random fields model, which is a typical method for modeling sequence labeling problems and it is frequently used for NER task in other languages. As in recent years deep learning has become a hot topic and it is widely used in many natural language processing tasks, we also create a system based on deep learning using LSTM neural networks. Our results indicate that using the proposed hybrid system (including statistical and list-based model), we can reach 84% in terms of F1 measure for seven labels: person, location, organization, date, time, percent and currency.

*: corresponding author

## 1 مقدمه

موجودیت‌های اسمی[1]، واحدهای اسمی معنادار متن هستند که مفهوم و منظور اصلی یک متن را مشخص می‌کنند. از مهم‌ترین انواع موجودیت‌های اسمی می‌توان به اسامی افراد، سازمان‌ها، مکان‌ها، پول، درصد، تاریخ، و زمان اشاره کرد. سیستم‌های تشخیص موجودیت اسمی، به شناسایی موجودیت‌های اسمی یک متن می‌پردازند و آن‌ها را در یکی از انواع مشخص طبقه‌بندی می‌کنند. تشخیص موجودیت‌های اسمی کاربردهای فراوانی در سیستم‌های استخراج اطلاعات، سیستم‌های پرسش و پاسخ، طبقه‌بندی متون، خلاصه سازی متون و بهینه سازی جستجو دارد.

اگرچه تحقیقاتی که در زمینه‌ی تشخیص موجودیت‌های اسمی در زبان‌های مختلف انجام شده وسیع بوده و بعضا نتایج قابل قبولی حاصل شده است، فعالیت‌های انجام‌شده در این زمینه در زبان فارسی چندان گسترده نبوده و نتایج ارایه شده پاسخ‌گوی نیازها نیست. در زبان فارسی، مجموعه داده‌ی استانداردی برای آموزش و ارزیابی سیستم‌های تشخیص موجودیت اسمی وجود ندارد که همین مسئله انجام پژوهش در این زمینه را بسیار دشوار کرده است. در این پژوهش به ساخت چنین مجموعه داده‌ای می‌پردازیم. سپس از این مجموعه داده برای آموزش و ارزیابی سیستمی جهت شناسایی و طبقه‌بندی موجودیت‌های اسمی در متون فارسی استفاده می‌کنیم.

در بخش ۲ به بررسی و مرور اجمالی سیستم‌های تشخیص موجودیت اسمی موجود برای زبان‌های انگلیسی و فارسی می‌پردازیم. سپس در بخش سوم چگونگی ساخت مجموعه داده‌ی استاندارد موجودیت‌های اسمی را به طور مشروح توضیح می‌دهیم. در بخش چهارم به شرح سیستم طراحی‌شده برای زبان فارسی در این پژوهش می‌پردازیم و در بخش پنجم آزمایش‌های انجام‌شده با استفاده از سیستم طراحی‌شده و مجموعه داده‌ی فراهم‌شده را شرح می‌دهیم و در نهایت در بخش ششم به جمع‌بندی و بیان کارهای آینده می‌پردازیم.

## 2 کارهای پیشین

با توجه به آن که در پی آن هستیم که مجموعه داده‌ی استانداردی برای زبان فارسی ایجاد کنیم، ابتدا به مروری گذرا روی مجموعه داده‌های موجود زبان انگلیسی می‌پردازیم و از آن جا که می‌خواهیم با بررسی کارهای پیشین انجام‌شده در زبان فارسی و با استفاده از روش‌های موجود در زبان انگلیسی به طراحی سیستمی کارآمد برای تشخیص موجودیت‌های اسمی زبان فارسی بپردازیم، در ادامه به مرور اجمالی پژوهش‌های انجام‌شده در زبان انگلیسی و سپس به بررسی پژوهش‌های زبان فارسی می‌پردازیم. همچنین از آن جا که انتخاب ویژگی‌ها به اندازه‌ی انتخاب روش در این مسئله اهمیت دارد، هر بخش را به دو زیربخش تقسیم می‌کنیم. در زیربخش اول به مرور روش‌های ارائه‌شده و در زیربخش دوم به معرفی ویژگی‌های ارائه‌شده برای حل مسئله‌ی تشخیص موجودیت‌های اسمی می‌پردازیم.

### ۲.۱ مجموعه داده‌ها

در [20] نشان داده شده است که سیستم تشخیص موجودیت‌های اسمی به حوزه و ژانر متن حساس است و سیستمی که برای یک حوزه (مثلاً حوزه‌ی پزشکی) و یا یک ژانر (مثل متون رسمی) آموزش داده شده، در سایر حوزه‌ها (مانند اجتماعی، سیاسی و ...) و یا سایر ژانرها (مثل وبلاگ یا میکروبلاگ) بیست تا چهل درصد کاهش کیفیت خواهد داشت. بنابراین به نظر می‌رسد اینکه سیستم جامعی آموزش داده شود که روی تمام حوزه‌ها و یا ژانرها عملکرد خوبی داشته باشد، عملی نیست. با این وجود انتخاب متون خبری از این حیث که دربرگیرنده‌ی حوزه‌های متعددی از متون در ژانر رسمی است، به نظر انتخاب خوبی می‌رسد و بررسی مجموعه داده‌های استاندارد موجود در زبان انگلیسی هم بر این ادعا صحه می‌گذارد. چرا که در تمامی مجموعه داده‌های استاندارد بررسی شده، مجموعه متون از متون خبری انتخاب شده‌اند.

در مجموعه داده‌ی CoNLL2003[2] که برای دو زبان انگلیسی و آلمانی ایجاد شده است، از مجموعه متون خبرگزاری رویترز برای زبان انگلیسی و مجموعه متون خبری روزنامه‌ی **Frankfurter Rundschau** برای زبان آلمانی استفاده شده است. این مجموعه داده از ۱۴۹۸۷ جمله که معادل ۲۰۳۶۲۱ کلمه است تشکیل شده است.

---

[1] named entities

[2] http://www.clips.uantwerpen.be/conll2003/ner/

در مجموعه داده‌ی MUC6[3] از مجموعه متون خبری وال استریت ژورنال و در مجموعه داده‌ی MUC7[4] از مجموعه خبرهای موجود در نیویورک‌تایمز استفاده شده است.

همچنین در مجموعه داده‌ی ACE2[5] از متون خبری واشنگتن‌پست، VOA و ...، در مجموعه داده‌ی ACE2003[6] از اخبار پخش‌شده، newswire ها و... و در مجموعه داده‌ی ACE2004[7] از متون اخبار پخش‌شده‌ی رادیویی استفاده شده است.

## ۲.۲ زبان انگلیسی

### 2.2.1 روش‌ها

مسئله تشخیص موجودیت‌های اسمی ذاتا از جنس مسائل برچسب‌گذاری دنباله است. منظور از برچسب‌گذاری دنباله، در اختیار داشتن دنباله‌ای از اشیاء است که با توجه به ترتیب آن‌ها قرار است به تک تک اشیاء برچسبی تعلق گیرد. در مسئله‌ی تشخیص موجودیت‌های اسمی این اشیاء، کلمات و برچسب‌ها، انواع موجودیت‌های اسمی هستند. به همین دلیل از ابتدای پژوهش در این مسئله، از روش‌های معروف و فراگیر برچسب‌گذاری دنباله استفاده شده است. در یک تلاش ابتدایی در [5] سیستمی با استفاده از زنجیره‌ی پنهان مارکف[8] ارائه شد. در این سیستم از روش زنجیره‌ی پنهان مارکف مرتبه اول که از معروف‌ترین و متداول‌ترین روش‌های معرفی‌شده برای برچسب‌گذاری دنباله‌هاست، استفاده شده است. در مدل زنجیره‌ی پنهان مارکف، یک سری متغیر مشاهده‌شده وجود دارد که با توجه به آن‌ها و با دانستن احتمال انتقال بین حالت‌ها به دنبال دنباله‌ای از حالت‌های پنهان با بیشترین احتمال هستیم. در مسئله‌ی تشخیص موجودیت‌های اسمی، متغیرهای مشاهده‌شده، کلمات و حالت‌های پنهان برچسب‌ها هستند.

در [6] از مدل بیشینه‌ی بی‌نظمی مارکف[9] استفاده شده است. در این روش (MEMM) بر خلاف زنجیره‌ی پنهان مارکف که تنها از برچسب متغیر قبلی برای تعیین برچسب متغیر کنونی استفاده می‌کند، امکان استفاده از ویژگی‌های دیگر نیز وجود دارد.

در [16] از روش میدان‌های تصادفی شرطی برای تشخیص موجودیت‌های اسمی استفاده شده است. در این روش، تمامی برچسب‌های خروجی با درنظر گرفتن کل ورودی تولید می‌شود. در اکثر مقالاتی که پس از این مقاله برای حل مسئله‌ی تشخیص موجودیت‌های اسمی ارائه‌شده‌اند و مورد مطالعه قرار گرفته‌اند، از روش CRF به عنوان روش اصلی و پایه‌ای استفاده شده است.

در هر سه روش مطرح شده از فرض ساده‌سازی مارکف برای تصمیم‌گیری در مورد برچسب‌ها استفاده شده است. فرض مارکف بیان می‌کند که برچسب هر متغیر تنها به چند متغیر در یک پنجره‌ی محدود اطراف آن بستگی دارد. اگر از این فرض استفاده نشود، محاسبات و تعداد متغیرها به قدری زیاد می‌شود که راه بهینه‌ای برای حل مسئله وجود نخواهد داشت. در [10] از روش CRF برای حل مسئله‌ی تشخیص موجودیت‌های اسمی استفاده شده است. ولی برای آن که بتوان از درجه‌های بالاتر وابستگی بین متغیرها استفاده کرد، از روش نمونه‌برداری گیبس استفاده کرده است.

در اغلب مقالات از روش میدان‌های تصادفی شرطی استفاده شده است. ولی در [22] از روش رده‌بندی بردارهای پشتیبان (SVC) استفاده شده است.

در سال‌های اخیر با ظهور شبکه‌های عصبی عمیق تحول شگرفی در روش‌های ارائه‌شده برای مسائل پردازش زبان طبیعی رخ داده است. برای مسئله‌ی تشخیص موجودیت‌های اسمی نیز تعدادی پژوهش بر این اساس صورت گرفته است. در [9] از نمایش طیفی کاراکترها[10] که از طریق اعمال یک شبکه عصبی هم‌گشتی[11] به دست آمده است، استفاده شده است. این سیستم مستقل از زبان است و نیازی به تعریف ویژگی ندارد و ویژگی‌ها را به صورت خودکار یادگیری می‌کند.

---

[3] https://catalog.ldc.upenn.edu/LDC2003T13
[4] https://catalog.ldc.upenn.edu/LDC2001T02
[5] https://catalog.ldc.upenn.edu/LDC2003T11
[6] https://catalog.ldc.upenn.edu/LDC2004T09
[7] https://catalog.ldc.upenn.edu/LDC2005T09
[8] Hidden Markov Model (HMM)
[9] Maximum Entropy Markov Model
[10] character embedding
[11] Convolutional Neural Network (CNN)

در [8] از یک سیستم ترکیبی از شبکه‌های عصبی هم‌گشتی برای یادگیری نمایش توزیع‌شده‌ی کاراکترها و شبکه‌های عصبی بازگشتی[12]، از نوع **LSTM** برای حفظ وابستگی‌های طولانی بین کلمات استفاده شده است.

در [13] روشی ترکیبی از شبکه‌های عصبی بازگشتی و میدان‌های شرطی تصادفی استفاده شده است. با استفاده از یک شبکه‌ی عصبی بازگشتی از نوع **LSTM**، زمینه‌ی سمت چپ و راست کلمات مدل می‌شود. سپس از ترکیب آن‌ها زمینه‌ی کلی کلمه به دست می‌آید. در نهایت به جای نمایش اولیه‌ی عبارات، الگوریتم میدان‌های شرطی تصادفی روی نمایش حاصل از این ترکیب اعمال می‌شود.

## 2.2.2 ویژگی‌ها

در[13] مجموعه‌ی خوب و جامعی از ویژگی‌ها پیشنهاد شده است. ویژگی‌های اصلی پیشنهادشده را می‌توان به دو دسته‌ی پایه و ریشه‌یابی‌شده تقسیم کرد که ویژگی‌های ریشه‌یابی‌شده همان ویژگی‌های پایه هستند که به جای خود کلمات به ریشه‌ی آن‌ها اعمال می‌شود.

این مقاله ویژگی‌های پایه را به ۶ دسته‌ی کلمه (نمایش برداری کلمه‌ی جاری و ۲ کلمه قبل و بعد از آن در ۵ بردار جداگانه)، کیف کلمات (نمایش برداری کلمه‌ی جاری و ۲ کلمه قبل و بعد از آن در یک بردار بدون نگه داشتن اطلاعات مربوط به مکان قرارگیری)، ان‌گرام (مشابه ویژگی کلمه ولی به جای کلمه (یونیگرام)، بایگرام جایگزین می‌شود)، ویژگی‌های املایی (بزرگ بودن تمام حروف کلمه، بزرگ بودن تنها حرف اول کلمه، ترکیبی از حروف بزرگ و کوچک، شامل بودن عدد، شامل بودن علائمی مانند '،' ، '.' و '...' ،'،' ، مخفف بودن و ... )، الگوهای املایی (مشابه ویژگی کلمه، ولی با این تفاوت که کلمات یکسان‌سازی می‌شوند. به این ترتیب که تمام حروف بزرگ به 'A'، حروف کوچک به 'a'، ارقام به '1'، فاصله‌ها به یک ' ' و سایر علائم به '-' تغییر یابند.) و وندها (پیشوند و پسوند کلمه شامل ۴-۲ حرف اول یا آخر کلمه) تقسیم کرده است.

در [19] ویژگی‌ها به سه دسته تقسیم شده‌اند که در ادامه به توضیح آن‌ها می‌پردازیم.

۱- ویژگی‌های سطح کلمه: بر اساس کاراکترهای کلمه تعریف می‌شوند و شامل ویژگی‌های املایی که گفته شد نیز می‌شوند. از جمله‌ی این ویژگی‌ها می‌تواند به بزرگ و کوچکی حروف، علائم نگارشی، ساخت‌واژه (پیشوند، پسوند و ریشه‌ی کلمات)، برچسب ادات سخن، و طول کلمه اشاره کرد.

۲- ویژگی‌های مبتنی بر دیکشنری یا لیستی: یکی از کارهایی که به سیستم تشخیص موجودیت‌های اسمی کمک می‌کند، استفاده از لیست‌هایی از اسامی خاص اشخاص مشهور، سازمان‌ها، مکان‌ها و مخفف‌های آن‌هاست. ویژگی‌های مبتنی بر لیست بر همین اساس تعریف می‌شوند، به این صورت که حضور یا عدم حضور آن‌ها در لیست‌های هر یک از موجودیت‌ها به عنوان یک ویژگی در نظرگرفته می‌شود.

۳- ویژگی‌های سطح سند و پیکره: این ویژگی‌ها بر اساس محتوا و ساختار سند تعریف می‌شوند. از جمله‌ی این دسته از ویژگی‌ها می‌توان به تعداد رخدادهای کلمه در سند، تعداد رخدادهای کلمه با حروف کوچک و بزرگ در سند، و محل قرارگیری در جمله اشاره کرد.

در[23] تأثیر روش‌های نمایش کلمات به عنوان ویژگی بررسی شده است. در این مقاله از روش خوشه‌بندی براون[13] و روش‌های نمایش طیفی کلمات **HLBL** و **Collobert and Weston** استفاده شده است. با بررسی روش‌های مختلف نمایش کلمات و اضافه کردن آن‌ها به عنوان ویژگی به یک مدل حاضر، به این نتیجه رسیده است که استفاده از خوشه‌بندی براون در مسئله‌ی تشخیص موجودیت‌های اسمی، موثرتر از استفاده از روش‌های نمایش توزیع‌شده‌ی مذکور است. در این روش کلمات به صورت سلسله‌مراتبی خوشه‌بندی می‌شوند.

در [18] از این روش برای رده‌بندی کلمات در کلاس‌های مختلف موجودیت‌های اسمی استفاده شده است. به این ترتیب با دانستن اینکه به عنوان مثال **Microsoft** برچسب سازمان می‌گیرد، از این واقعیت که **NIKE** با **Microsoft** در یک خوشه قرار گرفته استفاده کرده و به **NIKE** هم برچسب سازمان تخصیص می‌دهد.

در [21] نیز روش‌های مختلف نمایش توزیع‌شده برای افزودن به عنوان ویژگی با هم مقایسه شده‌اند که روش word2vec ارائه‌شده در [17] مناسب‌تر از سایرین بوده است. در [22] هم از **word2vec** استفاده شده است، ولی

---

[12] Recurrent Neural Network (RNN)
[13] Brown clustering

به جای آن که نمایش توزیع‌شده به صورت مستقیم به عنوان ویژگی به کار رود، نمایش‌های توزیع‌شده خوشه‌بندی شده و شماره‌ی خوشه‌ی هر کلمه به عنوان ویژگی به آن اضافه شده است. به این ترتیب مقدار این ویژگی برای کلمات با معانی یا کاربردهای نزدیک به هم مشابه خواهد بود.

## ۲.۳ زبان فارسی

### 2.3.1 روش‌ها

در [3]به معرفی سیستمی جهت تشخیص و دسته‌بندی موجودیت‌های اسمی در زبان فارسی پرداخته شده است. این سیستم با به‌کارگیری الگوهای متنی ممکن برای اسم‌های خاص متعلق به یک دسته، سعی در شناسایی موجودیت‌های اسمی دارد و از برچسب‌های نحوی و معنایی برای رفع ابهام استفاده می‌کند. این سیستم به طور کامل قاعده‌محور است و از روش‌های یادگیری استفاده نمی‌کند.

در [1]با استفاده از مجموعه داده‌ی پژوهشکده‌ی هوشمند علائم، روش‌های گوناگونی شامل روش شبکه عصبی، طبقه‌بند K نزدیک‌ترین همسایه، طبقه‌بند خطی و طبقه‌بند بیزین مورد آزمون قرار گرفته است و نشان داده شده است که طبقه‌بند خطی بهترین و طبقه‌بند مبتنی بر شبکه عصبی بدترین نتیجه را از نظر معیار **F1** نتیجه می‌دهند.

در [4]با استفاده از ترکیب مدل مخفی مارکف و قواعد تعیین شده، سیستمی برای تشخیص موجودیت‌های اسمی پیشنهاد شده است. در واقع این سیستم از نوع ترکیبی است (ترکیب مبتنی بر قاعده و یادگیری ماشین). آن‌ها سیستم خود را روی مجموعه داده‌ی برچسب‌خورده‌ای که از متون خبرگزاری مهر ساخته‌اند، مورد آزمون قرار داده‌اند.

در [2]، روشی ترکیبی از روش‌های مبتنی بر قاعده و آماری ارائه شده است. برای تهیه‌ی فهرست واژگان از ویکی‌پدیای فارسی استفاده شده است. همچنین به جز سیستم با سه نوع برچسب موجودیت اسمی (شخص، مکان و سازمان) از سیستم با ۶ نوع برچسب نیز استفاده شده است (شخص، مکان، سازمان، امکانات، محصول و رویداد). داده‌های آزمون به صورت دستی و با استفاده از پیکره‌ی بیجن‌خان تهیه شده است. همچنین برای فراهم کردن یک سیستم تشخیص موجودیت‌های اسمی مبتنی بر یادگیری ماشین، به صورت دستی داده‌های آموزش فراهم شده است و روش‌های یادگیری ماشین مورد آزمون قرار گرفته که بهترین مدل ارائه شده مدل مبتنی بر میدان‌های تصادفی شرطی بوده است.

### 2.3.2 ویژگی‌ها

در زبان انگلیسی ویژگی‌ای که بسیار مورد استفاده قرار گرفته و تاثیر به‌سزایی بر دقت سیستم‌های تشخیص موجودیت اسمی داشته، ویژگی بزرگ و کوچکی حروف است که در زبان فارسی موجود نیست. به جز آن، در زبان انگلیسی معمولا کلمات مخفف که عمدتا نشان‌دهنده‌ی نام سازمان‌ها هستند، با حروف تمام‌بزرگ و یا با نقطه در میان حروف مشخص می‌شوند و قابل تشخیص هستند، ولی در فارسی چنین نیست. به عنوان مثال '**CIA**' در زبان انگلیسی به دلیل بزرگ بودن تمام حروف کاندیدای خوبی برای نام سازمان است، ولی در زبان فارسی «ناجا» به این ترتیب قابل تشخیص نیست.

در عوض در زبان فارسی ویژگی کسره‌ی اضافه وجود دارد که می‌تواند مشخص‌کننده‌ی خوبی برای مرز عبارت‌های موجودیت اسمی باشد. به عنوان مثال پایان عبارت اسمی «سازمانِ شهرداریِ استانِ تهران» که باید به عنوان موجودیت اسمی برچسب سازمان بگیرد از این طریق قابل تشخیص است. تاثیر ویژگی کسره‌ی اضافه در[2]مورد بررسی قرار گرفته است و بهبود ۴ درصدی در حالت سه برچسبی شخص، مکان و سازمان نسبت به حالت عدم استفاده از این ویژگی گزارش شده است.

در [12] از ویژگی‌های مبتنی بر سند (جایگاه کلمه در جمله و سند) و مبتنی بر لیست (حضور یا عدم حضور در لیست‌های خاص انواع مختلف موجودیت‌های اسمی) استفاده شده است. همچنین خروجی سیستم مبتنی بر قاعده به عنوان یک ویژگی به سیستم مبتنی بر یادگیری ماشین داده شده است.

در [1]از نقش دستوری کلمه‌ی قبل و بعد، طول کلمه، جمع یا مفرد بودن اسم، وجود پسوندهای خاص برخی موجودیت‌های اسمی، وجود «ی» نسبی در آخر کلمه، و درصد حضور کلمه در کل متون آموزشی به صورت اسم مکان و اسم خاص شخص استفاده شده است.

# ۳ پیکره

در این بخش شرح فرآیند تولید پیکره موجودیت‌های اسمی ارائه می‌گردد. شرح مختصر شیوه‌نامه برچسب‌زنی، انتخاب اسناد پیکره، کیفیت‌سنجی برچسب‌زنی و آمار پیکره نهایی در ادامه آورده می‌شود.

## ۳.۱ شیوه‌نامه

تهیه شیوه‌نامه برچسب‌زنی موجودیت‌های اسمی که حاوی قواعد تشخیص و برچسب‌زنی موجودیت‌های مختلف سازمان، شخص، مکان، تاریخ، پول، و درصد است با رجوع به دو شیوه‌نامه‌ی استاندارد MUC[14] و CoNLL[15] و با تطبیق با قواعد زبان فارسی تهیه می‌گردد. برچسب‌زن با توجه به بافت، کلمات یا عباراتی را که جزء موجودیت‌های هفت‌گانه پیش‌گفته دسته‌بندی می‌شوند تعیین می‌نماید. شیوه کدگذاری برچسب موجودیت‌ها IOB یعنی برچسب اولین توکن (کلمه) با B، و برچسب توکن‌های (کلمات) دیگر، در صورت وجود، با I شروع می‌شود. کلماتی نیز که موجودیت نیستند برچسب O خواهند داشت.

قواعد کلی برچسب‌زنی تایید می‌نماید که موجودیت‌هایی که در پی هم می‌آیند تا حد امکان برچسب جداگانه می‌گیرند. دنباله عباراتی که جهت تعیین یک موجودیت واحد آورده می‌شوند، مانند «دانشکده برق و کامپیوتر دانشگاه تهران» همگی یک برچسب موجودیت می‌گیرند. در دلالت‌های التزامی که معمولا بین موجودیت مکان و سازمان رخ می‌دهد و نیاز به مرجع‌یابی وجود دارد، مرجع‌یابی انجام نمی‌شود. به عنوان مثال در «ایران ادعای آمریکا را رد کرد» کلمات «ایران» و «آمریکا» برچسب مکان می‌گیرند.

در عبارات هم‌پایه همه عبارت یک برچسب می‌گیرد. به عنوان نمونه «آمریکای شمالی و جنوبی» کلا یک برچسب مکان می‌گیرد. هم‌نامی‌ها نیز جزء موجودیت لحاظ می‌شوند و برچسب می‌گیرند. به عنوان مثال «فرودگاه بین‌المللی تهران» که هم‌نام «فرودگاه بین‌المللی امام خمینی» است برچسب موجودیت می‌گیرد. اگر صفات درون عبارت موجودیت قرار گیرد برچسب می‌گیرد، ولی اگر خارج از عبارت موجودیت باشد برچسب نمی‌گیرد. به عنوان نمونه «خلیج همیشگی فارس» کلا یک برچسب مکان می‌گیرد، ولی در «سید حسین محلاتی جلیل‌القدر» تنها بخش «سید حسین محلاتی» برچسب شخص می‌گیرد.

آنچه گفته شد مختصری از شیوه‌نامه برچسب‌زنی موجودیت‌های اسمی بود. به جهت رعایت اختصار ذکر جزئیات شیوه‌نامه برچسب‌زنی در همه موجودیت‌ها مقدور نیست. به همین دلیل به ذکر کلیات بسنده می‌کنیم و متن کامل شیوه‌نامه را با پیکره موجودیت‌های اسمی منتشر خواهیم نمود.

## ۳.۲ انتخاب اسناد

با هدف افزایش تنوع اسناد پیکره، منابع دریافت اخبار فهرستی از خبرگزاری‌ها و سایت‌های خبری مختلف در نظر گرفته می‌شود: خبرگزاری فارس، خبرگزاری مهر، ایرنا (خبرگزاری جمهوری اسلامی)، ایسنا (خبرگزاری دانشجویان ایران)، همشهری آنلاین، تابناک، فرارو، ورزش سه، انتخاب، و باشگاه خبرنگاران جوان.

انتخاب اسناد مجموعه داده از بین تعداد بسیار زیاد اخبار باید به گونه‌ای انجام شود که توزیع موضوعی اسناد پیکره نهایی با توزیع موضوعی اخبار دنیای واقعی هم‌خوانی داشته باشد و میانگین طول اسناد مجموعه داده نیز به میانگین طول اسناد دنیای واقعی نزدیک باشد. همچنین برای اینکه تنوع بیشتری در اسناد مجموعه داده نهایی وجود داشته باشد باید اسناد از زمان‌های مختلف انتخاب شوند. با این ترتیب رفتار اسناد پیکره به اخبار دنیای واقعی نزدیک‌تر می‌شود.

برای انتخاب اسناد مجموعه داده ابتدا نیاز است اسناد خبری از منابع ذکرشده با بکارگیری یک خزش‌گر جمع‌آوری شود. بازه زمانی دریافت اخبار ده ماه اول سال ۹۵ در نظر گرفته شده است. پس از حذف اسناد نامناسب که تنها شامل تصاویر یا متن‌های بسیار کوتاه هستند (زیر ۷۰ کاراکتر) در مجموع حدود ۷۰۰هزار سند خبری به دست می‌آید. پیش

---

پردازش اسناد که شامل نرمال‌سازی و واحدسازی متن است با ابزار Persianp[16] صورت گرفته است.

برای تعیین موضوع اسناد از سیستم رده‌بند موضوعی اسناد خبری پروژه هشتگ فارسی استفاده می‌شود و اسناد در شش دسته موضوعی "سیاست"، "اقتصاد"، "ورزش"، "فرهنگ و هنر"، "دانش و فناوری" و "جامعه" دسته‌بندی می‌شوند.

در هر موضوع میانگین طول و انحراف معیار اسناد محاسبه و با توجه به این دو پارامتر از اسناد طبق توزیع نرمال نمونه‌گیری انجام می‌شود. با این روش میانگین طول اسناد با دنیای واقعی یکسان و پراکندگی اسناد نسبت به پارامتر طول نیز حفظ می‌گردد. سپس در بازه‌های زمانی مختلف به صورت میانگین از موضوعات مختلف به گونه‌ای سند انتخاب می‌شود که توزیع موضوعی اسناد انتخابی با توزیع موضوعی همه اسناد برابر باشد.

برای اینکه پیکره در نهایت شامل اسنادی از منابع مختلف خبری باشد، از هر منبع خبری به نسبت اخباری که هر منبع خبری در یک موضوع منتشر کرده است از مجموع اسناد نمونه‌گیری شده سند انتخاب می‌شود. چون برخی از خبرگزاری‌ها اخبار بسیار بیشتری نسبت به دیگر خبرگزاری‌ها و سایت‌های خبری منتشر می‌کنند و جهت رعایت توازن، حداکثر ۲۰ درصد اسناد مجموعه داده نهایی می‌تواند از یک منبع خبری باشد. تنها مورد استثنا سایت ورزش سه است که چون اختصاص به ورزش دارد ۳۰ درصد اسناد ورزشی را به خود اختصاص می‌دهد.

به این ترتیب پیکره نهایی شامل اسنادی است که از نظر موضوع، میانگین طول و زمان با اخبار دنیای واقع هم‌خوانی دارد. شکل ۱ نمودار تعداد اسناد را در موضوعات مختلف نشان می‌دهد.

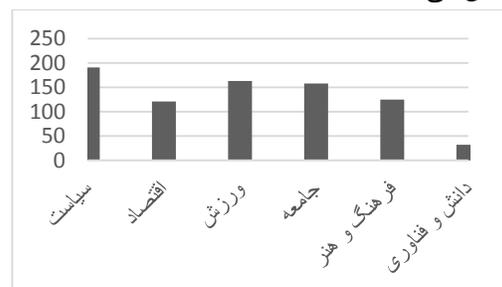

**شکل ۱ نمودار تعداد اسناد انتخابی در موضوعات مختلف**
**Figure 1. Number of chosen documents for each news category**

## ۳.۳ کیفیت برچسب‌زنی

برای اینکه برچسب‌زنی کیفیت‌سنجی شود، بخشی از پیکره توسط دو شخص مختلف برچسب‌گذاری می‌شود و میزان توافق بین برچسب‌ها محاسبه می‌گردد. از اسناد مجموعه داده ۱۶۰ سند برچسب‌خورده که در مجموعه حدود ۲۰۳۰ جمله دارد برای برچسب‌زنی مجدد انتخاب می‌شود. این اسناد شامل ۵۳۸۰۰ توکن هستند. اگر اختلافی بین برچسب‌ها مشاهده شود، یا معلول اختلاف برداشت از بافت کلمه و تشخیص متفاوت در نوع موجودیت است یا اشتباهی توسط یکی از برچسب‌زن‌ها رخ داده و نوع برچسب اشتباه وارد شده است. در این ۱۶۰ سند تعداد اختلاف بین دو برچسب‌زن در همه توکن‌ها ۳۸۶ مورد است و میزان هماهنگی برچسب‌زن‌ها ۹۹ درصد است.

با توجه به اینکه درصد بالایی از توکن‌ها موجودیت اسمی نیستند و برچسب «O» دارند، می‌توان تنها با لحاظ کردن توکن‌هایی که برچسب موجودیت دارند نیز درصد اختلاف را بررسی کرد. چون هیچ مرجع مطلقی برای تعیین اینکه کدام توکن‌ها موجودیت اسمی هستند وجود ندارد، برای محاسبه تعداد توکن‌هایی که موجودیت اسمی هستند اجتماع تعداد توکن‌هایی که توسط برچسب‌زن‌ها موجودیت تشخیص داده شده‌اند ملاک قرار می‌گیرد. در مجموع ۷۶۸۳ توکن موجودیت اسمی دارند. با در نظر گرفتن تعداد اختلافات (۳۸۶) در این حالات میزان توافق برچسب‌زن‌ها ۹۵ درصد است.

همچنین معیار **Kappa** برای محاسبه میزان توافق بین برچسب‌زن‌ها را محاسبه می‌کنیم. معیار **Kappa** توافق را بر اساس میزان فاصله رای برچسب‌زن‌ها نسبت به حالتی که برچسب‌زنی را تصادفی انجام دهند محاسبه می‌کند. نتایج در حالتی که همه برچسب‌ها من‌جمله برچسب «O» وجود دارد به شرح زیر است:

**p-value < 2.2e-16 ;**
**95percent confidence interval: 0.9695260**
**0.9750365**
**sample estimates:0.9722813**

نتایج در حالتی که برچسب «O» در نظر گرفته نمی‌شود به شرح زیر است:

**p-value < 2.2e-16 ;**
**95percent confidence interval: 0.9351699**
**0.9466587**
**sample estimates: 0.9409143**

---

[16] persianp.ir

همانطور که مشاهده می‌شود میزان p-value در هر دو حالت بسیار پایین است و نتیجه حکایت از توافق بالا بین برچسب‌زن‌ها دارد.

## 3.4 آمار پیکره

پیکره نهایی شامل 709 سند خبری است که در مجموع شامل 7145 جمله است. در این مجموعه داده 302530 توکن (کلمه) وجود دارد که 41148 توکن برچسب موجودیت دارند و مابقی توکن‌ها برچسب موجودیت ندارند (برچسب آن‌ها O است). آمار تعداد موجودیت‌های مختلف (که تک‌کلمات یا عباراتی هستند که یک موجودیت را نشان می‌دهند) و تعداد توکن‌های برچسب‌خورده در هر کلاس موجودیت در جدول 1 ذکر شده است.

**جدول 1 تعداد توکن‌های برچسب‌خورده در هر کلاس موجودیت**
**Table 1. Number of annotated tokens for each entity type**

| تعداد موجودیت‌های یکتا | تعداد توکن‌ها | تعداد موجودیت | نوع موجودیت |
|---|---|---|---|
| 2507 | 7675 | 4447 | شخص |
| 2413 | 16964 | 6360 | سازمان |
| 1288 | 8782 | 6223 | مکان |
| 801 | 732 | 281 | زمان |
| 157 | 4259 | 1858 | تاریخ |
| 436 | 2037 | 527 | مبالغ مالی |
| 179 | 699 | 326 | درصد |
| 7781 | 41148 | 20022 | تعداد کل |

# 4 سیستم طراحی‌شده

برای پیاده‌سازی سیستم از روش‌های مبتنی برقاعده، آماری و یادگیری عمیق استفاده شده است که در ادامه به توضیح جزئیات هر یک می‌پردازیم.

## 4.1 روش قاعده‌محور

در روش‌های قاعده‌محور، قواعدی به صورت دستی بر اساس شمّ زبانی خبره یا به صورت نیمه‌خودکار و با مشاهده مصادیق متعدد تدوین می‌گردد که با استفاده از آن قواعد موجودیت‌های اسمی متن تشخیص داده شوند. قواعد به دو گونه قواعد منظم روی دنباله کاراکترهای هر توکن و قواعد منظم روی دنباله توکن‌های متن است. به عنوان مثال موجودیت تاریخ (مانند 12/2/1396) را می‌توان با استفاده از عبارات منظم روی دنباله کاراکترهای توکن تشخیص داد. قواعد منظم روی دنباله توکن‌ها نیز قابل تعریف هستند. این قواعد را می‌توان به صورت قواعد ساده پیش‌دنباله و فهرست (مانند قاعده «"دانشگاه آزاد اسلامی" + نام شهر» برای تشخیص موجودیت سازمان) یا قواعدی مشابه عبارات منظم ولی برای دنباله توکن‌ها تهیه و برای تشخیص موجودیت‌ها استفاده کرد. به عنوان مثال قاعده «(/آقای|خانم|آقایان)({tag:N})[+]» که دنباله توکن‌هایی که با یکی از کلمات «آقای»، «خانم» و «آقایان» شروع می‌شود و تعدادی توکن برچسب اسم (N) دارند تشخیص می‌دهد برای تشخیص اسم افراد استفاده می‌شود.

## 4.2 روش آماری

با توجه به آن چه در بخش قبل گفته شد، در بیشتر مقالات از روش CRF به دلیل تناسب کامل با نوع مسئله به عنوان روش اصلی و پایه در مسئله‌ی تشخیص موجودیت‌های اسمی استفاده شده است. حتی در روش‌های جدیدتر که در سال‌های اخیر با استفاده از یادگیری عمیق ارائه شده، کماکان از CRF در ترکیب با یادگیری عمیق استفاده شده است. به عنوان مثال در [15] از ترکیب CRF و شبکه‌های LSTM استفاده شده است. به همین دلیل تصمیم گرفتیم از این روش به عنوان روش آماری منتخب استفاده کنیم و تمرکز را در ادامه بر تعریف ویژگی‌های مناسب قرار دهیم. در ادامه ابتدا به توضیح مختصر روش میدان‌های شرطی تصادفی می‌پردازیم و سپس ویژگی‌های مورد استفاده را معرفی می‌کنیم.

### 4.2.1 میدان‌های شرطی تصادفی

روش بیشینه‌ی بی‌نظمی مارکف اگرچه قوت و دقت زیادی دارد و برای مدل کردن مسائل برچسب‌گذاری دنباله بسیار مناسب است، یک ضعف جدی دارد که از آن با عنوان

اریب برچسب یاد می‌شود. این مشکل را می‌توان این‌طور توضیح داد که حالت‌های با بی‌نظمی کم روی توزیع انتقال‌ها، در واقع مشاهده‌ی خود را نادیده می‌گیرند. برای حل این مشکل، با حفظ تمام مزایای روش MEMM، روش میدان‌های تصادفی شرطی (CRF) ارائه شده است. در شکل2 تفاوت روش HMM،MEMM و CRF در قالب تصویر نشان داده شده است. همان‌طور که در این شکل مشخص است، در روش HMM هر حالت تنها با توجه به حالت قبلی تولید می‌شود و مشاهده‌ها از حالت‌ها تولید می‌شوند. در MEMM، هر حالت با توجه به حالت قبلی و مشاهده‌ی کنونی تعیین می‌شود. در CRF، کل دنباله‌ی حالت‌ها از کل دنباله‌ی مشاهدات تولید می‌شود.

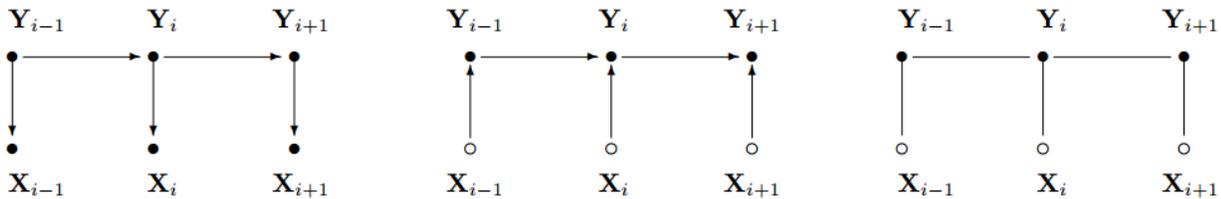

شکل2 تفاوت روشهای HMM،MEMM و CRF. برگرفته از [14] . در شکل سمت چپ HMM، در وسط MEMM و در سمت راست CRF نمایش داده شده است.
Figure 2. Difference between HMM, MEMM and CRF methods. (from left to right)

### 4.2.2 ویژگی‌ها

ویژگی‌هایی که ما برای آموزش سیستم CRF استفاده کرده‌ایم، ویژگی‌های ابتدایی و پایه‌ای هستند. ویژگی‌های املایی تعریف‌شده در مقالات انگلیسی به صورت مستقیم برای فارسی قابل استفاده نیست و باید برای زبانی فارسی بازتعریف شوند. برای مثال می‌توان به ویژگی دارا بودن ارقام به عنوان یک پیش‌فرض برای تعلق کلمه به دسته‌ی موجودیت‌های اسمی درصد، تاریخ و زمان اشاره کرد. ما از ویژگی‌های املایی استفاده نکرده‌ایم. همچنین به جای استفاده‌ی مستقیم از ویژگی کسره‌ی اضافه، از ویژگی برچسب عبارت اسمی[17] استفاده کردیم، چون موجودیت‌های اسمی معمولا یک عبارت اسمی کامل هستند. برچسب‌زنی ویژگی عبارت اسمی می‌تواند به صورت IOB انجام شود،یعنی کلمه اول موجودیت اسمی برچسب B، کلمات دیگر در موجودیت برچسب I و کلماتی که در عبارت اسمی نیستند برچسب O داشته باشند.

علاوه بر این، از ویژگی برچسب ادات سخن استفاده کردیم. ضمنا اگرچه استفاده از ریشه[18]ی کلمات در پژوهش‌های بازیابی اطلاعات مفید است، در سایر کارهای مربوط به پردازش زبان استفاده از بنواژه[19] نتیجه‌ی بهتری دارد [11]، زیرا در تعیین ریشه‌ی کلمات تنها به ظاهر کلمات توجه می‌شود در حالی که در تعیین بنواژه به سطح معنایی کلمات و ارتباط کلمات مجاور با هم نیز توجه می‌شود. به همین دلیل از ویژگی بنواژه نیز علاوه بر خود کلمه استفاده کردیم. علاوه بر این‌ها استفاده از ویژگی‌های مبتنی بر دیکشنری نیز کاربردی است. به این ترتیب که فهرستی از موجودیت‌های اسمی بدون ابهام فراهم کردیم و حضور یا عدم حضور کلمه در این فهرست را به عنوان ویژگی در نظر گرفتیم. به این ترتیب ویژگی‌های مورد استفاده از این قرار هستند:

- کلمه‌ی جاری و یک کلمه قبل و یک کلمه بعد از آن
- برچسب ادات سخن کلمه جاری و کلمه قبل و کلمه بعد
- برچسب عبارت اسمی کلمه جاری و کلمه قبل و کلمه بعد
- بنواژه کلمه‌ی جاری و کلمه قبل و کلمه بعد

---

[17] NP-Chunk
[18] stem
[19] lemma

- ان-گرام‌های سطح کاراکتر کلمه تا سقف ۶ کاراکتر (فقط از ابتدا یا انتهای کلمه): برای بررسی پیشوندها و پسوندها
- ویژگی‌های مبتنی بر دیکشنری به صورت تطابق جزئی

## 4.3 یادگیری عمیق

همان‌طور که در بخش قبل گفته شد، در اغلب سیستم‌های طراحی‌شده با استفاده از یادگیری عمیق از شبکه‌های عصبی بازگشتی با استفاده از LSTM استفاده شده است. این شبکه‌های عصبی برای مدل‌سازی مسائل از جنس برچسب‌گذاری دنباله‌ها مناسب هستند و استفاده از LSTM امکان استفاده از اطلاعات دور از کلمه‌ی جاری را برای برچسب‌گذاری فراهم می‌کند. ما نیز در این جا از شبکه عصبی با استفاده از LSTM استفاده کرده‌ایم.

### 4.3.1 ویژگی‌ها

برای استفاده از این روش امکان آن وجود داشت که ویژگی همراه با کلمات به شبکه داده شود و یا به کلی فرآیند یادگیری به شبکه عصبی واگذار شود. ویژگی‌های مورد استفاده‌ی ما به این ترتیب است:

- بن‌واژه‌ی کلمه
- برچسب ادات سخن
- برچسب عبارت اسمی

به این ترتیب سیستم طراحی‌شده توسط ما از چندین قسمت شامل بخش قاعده‌محور (شامل لیست‌محور و مبتنی بر عبارات منظم)، آماری مبتنی بر CRF و یادگیری عمیق مبتنی بر LSTM تشکیل می‌شود. در ادامه با انجام آزمایش‌های متعدد با استفاده از مجموعه داده‌ی طراحی‌شده به ارزیابی هر بخش می‌پردازیم.

# ۵ ارزیابی

در این بخش ابتدا به معرفی ابزارهای مورد استفاده می‌پردازیم. سپس نتایج ارزیابی سیستم با استفاده از سیستم قاعده‌محور، سیستم آماری و شبکه عصبی را ارائه می‌دهیم. همچنین به ارزیابی نتایج ترکیب سیستم قاعده‌محور و آماری می‌پردازیم.

## ۵.۱ ابزارها

### 5.1.1 ابزار تشخیص موجودیت اسمی استنفورد

گروه پردازش زبان طبیعی دانشگاه استنفورد[20]، مجموعه‌ای از ابزارهای پردازش زبان را به صورت متن‌باز در اختیار پژوهشگران قرار داده‌اند. یکی از این ابزارها تشخیص موجودیت‌های اسمی[21] است. این ابزار به زبان جاوا نوشته شده است و مبتنی بر روش **CRF** است. همچنین مجموعه‌ای غنی از ویژگی‌ها به صورت پیش‌فرض برای آن تعریف شده‌اند و به سادگی قابل استفاده هستند. علاوه بر این در بسیاری از مقالات از همین سیستم به عنوان سیستم پایه برای پیاده‌سازی یک ابزار تشخیص موجودیت اسمی استفاده شده است و مقاله‌ی متناظر آن [10] ارجاعات زیادی دارد.

### 5.1.2 ابزار استخراج ویژگی‌های کلمات فارسی

برای استخراج ویژگی‌های بن‌واژه، برچسب عبارت اسمی، و برچسب ادات سخن از ابزار **Persianp** استفاده شده است.

### 5.1.3 ابزار شبکه‌عصبی

برای اجرای شبکه عصبی از ابزار openNMT[22] استفاده شده است. این ابزار یک فریم‌ورک عمومی برای پیاده‌سازی انواع مدل‌های یادگیری عمیق است که به ویژه تمرکزش روی مدل‌های دنباله به دنباله است. در این مدل‌ها، ورودی یک دنباله و خروجی نیز یک دنباله است که لزوما طول دنباله‌ی خروجی با دنباله‌ی ورودی یکسان نیست. ولی جز این، حالتی هم برای مدل‌های برچسب‌زنی دنباله دارد که در آن طول دنباله‌ی خروجی و ورودی الزاما یکسان است. ما در این جا از این حالت استفاده کرده‌ایم. در این حالت شبکه عصبی از ۲ لایه LSTM تشکیل شده است و طول بردارهای طیفی کلمات ۵۰۰ و تعداد گره‌های شبکه نیز ۵۰۰ است.

---

## ۵.۲ آزمایش‌ها

در این بخش به ارائه و بررسی نتایج آزمایش روی بخش‌های مختلف سیستم طراحی‌شده می‌پردازیم. برای تمامی آزمایش‌ها از شیوه‌ی **K-fold** با **k=5** استفاده شده است.

### ۵.۲.۱ روش قاعده‌محور

برای تشخیص قاعده‌محور موجودیت‌های اسمی فارسی از ابزار **TokenRegex** استنفورد[7] استفاده می‌کنیم و قواعد را مختص ابزار آماده‌سازی می‌نماییم.

نتیجه‌ی ارزیابی سیستم قاعده‌محور طراحی‌شده بر روی کل مجموعه دادگان، به تفکیک نوع برچسب درجدول 2 گزارش شده است. همان طور که انتظار می‌رفت دقت این سیستم نسبتا بالا ولی فراخوانی آن پایین است.

جدول 2 نتایج سیستم قاعده‌محور
Table 2. Results of our rule-based system

|   | ER | OC | RG | IM | AT | ON | CT | ل |
|---|---|---|---|---|---|---|---|---|
| قت | .86 | .80 | .75 | .92 | .86 | .99 | .98 | .81 |
| راخوانی | .44 | .69 | .54 | .52 | .33 | .75 | .92 | .55 |
| 1 | .58 | .74 | .63 | .66 | .48 | .85 | .95 | .65 |

### ۵.۲.۲ آماری

نتیجه‌ی ارزیابی سیستم قاعده‌محور طراحی‌شده بر روی کل مجموعه دادگان، به تفکیک نوع برچسب در جدول 3 گزارش شده است. همان طور که از این جدول مشخص است، با استفاده از سیستم آماری به دقت، فراخوانی و در نتیجه **F1** بالاتری دست پیدا کرده‌ایم. این مسئله نشان می‌دهد که به دلیل به قدر کافی بزرگ بودن حجم مجموعه داده، سیستم آماری موفق به یادگیری قواعد اصلی تشخیص موجودیت‌های اسمی شده است. تعداد توکن‌هایی که با برچسب زمان در مجموعه داده مشخص شده‌اند، کم‌تر از سایر انواع موجودیت‌های اسمی بوده است، در حالی که تنوع قوانین برای این موجودیت زیاد است. به همین دلیل فراخوانی برای این برچسب کم‌تر از سایر برچسب‌ها بوده است. در مورد درصد و واحد پول با آن که تعداد توکن‌های مشخص‌شده با این برچسب‌ها در مجموعه داده کم‌تر از سایر موجودیت‌ها (به جز زمان) بوده است، دقت و فراخوانی بالایی به دست آمده است. علت این امر به کم‌تر بودن تعداد و پیچیدگی قواعد تشخیصی برای این دو برچسب برمی‌گردد.

جدول 3 نتایج سیستم آماری
Table 3. Results of our statistical system

|   | ER | OC | RG | IM | AT | ON | CT | ل |
|---|---|---|---|---|---|---|---|---|
| قت | .93 | .90 | .93 | .91 | .89 | .98 | .98 | .92 |
| راخوانی | .73 | .76 | .75 | .62 | .79 | .84 | .91 | .76 |
| 1 | .82 | .83 | .83 | .74 | .84 | .90 | .95 | .83 |

### ۵.۲.۳ ترکیبی

برای ترکیب دو سیستم قاعده‌محور و آماری، دو روش را آزمودیم. اول این که اولویت را به سیستم قاعده‌محور بدهیم و دیگر این که سیستم آماری در اولویت قرار بگیرد. منظور از اولویت یک سیستم آن است که ابتدا برچسب‌زنی متن با استفاده از آن سیستم انجام شود و سپس برای کلماتی که جزء هیچ موجودیت اسمی تشخیص داده نشده‌اند، از نتیجه‌ی سیستم دیگر استفاده کنیم.

نتایج درجدول 4 و جدول 5 گزارش شده است. ترکیب با سیستم قاعده‌محور در هیچ یک از دو حالت مورد آزمون‌قرارگرفته، باعث بهبود نتایج سیستم آماری نشده است. به نظر می‌رسد که به دلیل مناسب بودن حجم مجموعه داده‌ی مورداستفاده سیستم آماری به خوبی قواعد را شناسایی کرده است و ترکیب با سیستم قاعده‌محور کمکی به آن نمی‌کند.

جدول 4 نتایج سیستم ترکیبی (اولویت با سیستم قاعده‌محور)
Table 4. Results of the hybrid model (Rule-based has the higher priority)

|   | ER | OC | RG | IM | AT | ON | CT | ل |
|---|---|---|---|---|---|---|---|---|
| قت | .87 | .80 | .79 | .90 | .87 | .98 | .98 | .83 |
| راخوانی | .86 | .79 | .79 | .64 | .81 | .89 | .94 | .79 |
| 1 | .81 | .79 | .79 | .74 | .84 | .94 | .96 | .81 |

جدول 5 نتایج سیستم ترکیبی (اولویت با سیستم آماری)
Table 5. Results of the hybrid model (Statistical has the higher priority)

|   | CT | ON | AT | IM | RG | OC | ER |   |
|---|---|---|---|---|---|---|---|---|
| .84 | .98 | .98 | .87 | .90 | .80 | .85 | .88 | دقت |
| .81 | .94 | .89 | .81 | .66 | .82 | .80 | .77 | فراخوانی |
| .82 | .96 | .94 | .84 | .76 | .81 | .82 | .82 | F1 |

سیستم قاعده‌محور خود از دو بخش لیست‌محور و مبتنی‌بر عبارات منظم تشکیل شده است. برای ترکیب دو سیستم یک بار ترکیب سیستم لیست‌محور را با سیستم آماری بررسی کردیم، به این ترتیب که ابتدا متون با سیستم آماری برچسب خورده و سپس کلماتی که در خروجی این سیستم جزء هیچ موجودیت اسمی تشخیص داده نشده‌اند، با استفاده از سیستم لیست‌محور برچسب‌گذاری شد. نتایج در جدول 6 گزارش شده است. در این حالت به دلیل استفاده از لیست موجودیت‌ها که شامل سه برچسب شخص، مکان و سازمان بوده است، فراخوانی این سه برچسب افزایش پیدا کرده است و این مسئله منجر به بهبود نتیجه‌ی کلی سیستم ترکیبی شده است. در این حالت بهترین نتیجه از سیستم طراحی‌شده به دست آمده است.

جدول 6 نتایج سیستم ترکیبی (آماری و لیست‌محور)
Table 6. Results of the hybrid model (Statistical and list-based)

|   | CT | ON | AT | IM | RG | OC | ER |   |
|---|---|---|---|---|---|---|---|---|
| .91 | .98 | .98 | .89 | .91 | .92 | .90 | .92 | دقت |
| .77 | .91 | .84 | .79 | .62 | .76 | .78 | .74 | فراخوانی |
| .84 | .95 | .90 | .84 | .74 | .83 | .83 | .82 | F1 |

### 5.2.4 یادگیری عمیق

از آن جا که الگوریتم‌های یادگیری عمیق برای آن که به خوبی عمل یادگیری را انجام دهند به حجم زیادتری از داده‌ی آموزش نسبت به حالت یادگیری ماشین معمولی دارند، آزمایش‌های این حالت را با استفاده از K-fold و با K=10 انجام دادیم. آزمایش‌ها دو بار انجام شده‌اند. بار اول بدون افزودن ویژگی و بار بعد با استفاده از ویژگی‌های گفته‌شده در بخش قبل که نتایج به ترتیب در جدول 7 و جدول 8 گزارش شده است.

جدول 7 نتایج سیستم مبتنی‌بر یادگیری عمیق بدون ویژگی
Table 7. Results of the deep-learning based system without using any features

|   | CT | ON | AT | IM | RG | OC | ER |   |
|---|---|---|---|---|---|---|---|---|
| .84 | .79 | .81 | .75 | .66 | .86 | .85 | .86 | دقت |
| .76 | .47 | .55 | .69 | .46 | .80 | .78 | .80 | فراخوانی |
| .80 | .58 | .66 | .72 | .54 | .83 | .81 | .83 | F1 |

جدول 8 نتایج سیستم مبتنی‌بر یادگیری عمیق با استفاده از ویژگی‌ها
Table 8. Results of the deep-learning based system using features

|   | CT | ON | AT | IM | RG | OC | ER |   |
|---|---|---|---|---|---|---|---|---|
| .87 | .75 | .83 | .79 | .79 | .87 | .87 | .91 | دقت |
| .81 | .61 | .69 | .74 | .72 | .82 | .83 | .87 | فراخوانی |
| .84 | .67 | .75 | .76 | .76 | .84 | .85 | .89 | F1 |

در این حالت بدون استفاده از ویژگی‌های پیشوند، پسوند، کلمه‌ی قبل و بعد، و برچسب ادات سخن کلمه‌ی قبل و بعد و تنها با استفاده از سه ویژگی لم، برچسب ادات سخن کلمه‌ی جاری و برچسب عبارت اسمی آن به F1 معادل سیستم آماری آموزش‌داده‌شده رسیده‌ایم. فراخوانی این سیستم بالاتر از سیستم آماری، ولی دقت آن پایین‌تر از سیستم آماری است. یادگیری عمیق به شدت متأثر از حجم مجموعه داده‌ی مورد استفاده است و هر چه حجم مجموعه داده بزرگ‌تر باشد، امکان آن که بتواند قواعد را به درستی یاد‌گیری کند، بیشتر می‌شود. به همین دلیل در اینجا با توجه به جدول نتایج مشخص می‌شود که دقت بالاتری برای سه برچسب شخص، مکان و سازمان که بیشترین تعداد موجودیت را در مجموعه داده‌ی ساخته‌شده به خود اختصاص داده‌اند، به دست آمده است و برای واحد پول و درصد که تشخیص آن‌ها به دلیل کم بودن تعداد قواعد و همچنین کم بودن ابهام ساده‌تر است و هر دو سیستم آماری و قاعده‌محور برای آن‌ها به دقت بالاتر از ۹۸ درصد رسیده‌اند، به دقت کمتری دست پیدا کرده است، زیرا این دو موجودیت تعداد کمی نمونه در مجموعه داده‌ی ساخته‌شده داشته‌اند.

## 5.2.5 تحلیل خطا

برای تحلیل نتایج و شناخت منابع خطای سیستم می‌توان به ماتریس confusion برچسب‌زنی سیستم مراجعه کرد. این ماتریس در شکل 3 نمایش داده شده است (برای ترسیم این ماتریس از ابزار scikit-learn[23] در زبان پایتون استفاده شده است).

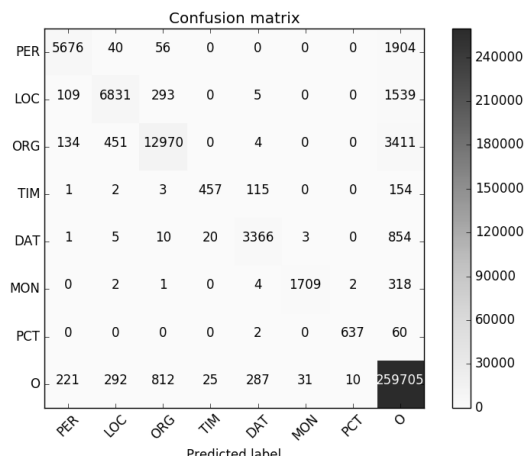

**شکل 3 ماتریس confusion سیستم تشخیص موجودیت اسمی (ستون‌ها برچسب‌های صحیح و ردیف‌ها برچسب‌های تشخیص‌داده شده است)**
**Figure 3. Confusion matrix of our NER system**

باتوجه به این ماتریس، منابع خطا چند دسته است. منبع اصلی خطا موجودیت‌های اسمی است که تشخیص داده نشده‌اند و یا کلمات عامی که به عنوان موجودیت اسمی تشخیص داده شده‌اند (ردیف آخر و اولین ستون از سمت راست). به عنوان مثال کلماتی مانند «هتل»، «شهرستان» و «جهان» در جاهایی که به تنهایی ظاهر شده‌اند برچسب مکان خورده‌اند یا کلمه‌ی «تیم» در جاهایی که به تنهایی ظاهر شده، برچسب سازمان گرفته است. در حالی که این اسامی در واقع اسم عام بوده‌اند و نباید برچسب موجودیت می‌گرفته‌اند. علت این خطاها حضور بالای این کلمات در موجودیت مکان است. در این حالات مدل آماری در تشخیص موجودیت بودن یا نبودن کلمه دچار خطا می‌شود.

همچنین بین سه موجودیت شخص، سازمان، و مکان تعداد خطاها بالاست. در مجموعه داده، تعداد توکن‌های این سه موجودیت از همه بیشتر است و میزان تداخل بین آن‌ها نیز بیشتر بوده است. در این قسمت تعداد بیشتر خطاها تداخل تشخیص بین موجودیت سازمان و موجودیت مکان است. به عنوان مثال وقتی از تیم‌های ورزشی با نام شهر آن‌ها صحبت شده است، سیستم آن‌ها را مکان تشخیص داده در حالی که در واقع موجودیت سازمان هستند. یا به طور عکس در جمله‌ی «آمریکاگردان زرهیکاملیراهیسرزمین‌هایشرقارپاخواهدکرد.»، «شرق اروپا» موجودیت مکان است که به اشتباه توسط سیستم به عنوان موجودیت سازمان برچسب‌گذاری شده است. همچنین بین دو برچسب شخص و مکان هم خطا وجود داشته است. به عنوان مثال در عبارت «به گزارش نسیم آنلاین»، «نسیم آنلاین» به علت وجود کلمه‌ی «نسیم» به عنوان موجودیت شخص تشخیص داده شده است، در حالی که در واقع سازمان بوده است. به جز این، بین دو موجودیت زمان و تاریخ نیز میزان خطا قابل توجه است.

به این ترتیب با توجه به ارزیابی‌های انجام‌گرفته در این بخش، بهترین سیستم طراحی‌شده‌ی ما ترکیب سیستم آماری و لیست‌محور بوده است. لیست‌های موجودیت‌ها در هر زمان قابل ویرایش و غنی‌سازی با موجودیت‌های جدیدتر مطرح‌شده است. تشخیص این سیستم با خطاهایی همراه است که باید به آن‌ها توجه کرد.

# ۶ نتیجه‌گیری و کارهای آینده

مسئله‌ی تشخیص موجودیت‌های اسمی به عنوان یک گام پیش‌پردازشی برای بسیاری از مسائل پردازش زبان طبیعی مطرح است. این مسئله در زبان فارسی به دلیل نبود یک مجموعه داده‌ی استاندارد، کمتر مورد پژوهش قرار گرفته است. در این پژوهش با بررسی ویژگی‌های مجموعه داده‌های استاندارد موجود در سایر زبان‌ها و به ویژه زبان انگلیسی، تلاش کردیم که مجموعه داده‌ی استانداردی برای زبان فارسی ایجاد کنیم. با توجه به این که در بسیاری از مجموعه داده‌های استاندارد موجود در زبان انگلیسی از متون خبری به عنوان منبع ساخت استفاده شده است، با جمع‌آوری متون خبری خبرگزاری‌های متعدد و با استفاده از برچسب‌گذاران انسانی، به ساخت مجموعه داده‌ی استانداردی برای زبان فارسی اقدام کردیم. سپس با مطالعه‌ی سیستم‌های طراحی‌شده برای تشخیص موجودیت‌های اسمی در زبان انگلیسی و با استفاده از مجموعه داده‌ی تهیه‌شده به طراحی

---

[23] http://scikit-learn.org/

سیستمی برای زبان فارسی پرداختیم. سیستم ما از بخش‌های متعدد شامل قاعده‌محور (تشکیل‌شده از لیست‌محور و قواعد منظم)، آماری و یادگیری عمیق تشکیل شده است. با انجام آزمایش‌هایی روی بخش‌های مختلف سیستم طراحی‌شده، به سیستم ترکیب‌شده از آماری و لیست‌محور به بهترین نتیجه بر اساس معیار F1 دست یافتیم.

با توجه به دقت و فراخوانی حاصل از ارزیابی روش‌های مختلف روی مجموعه داده‌ی فراهم‌شده و با مقایسه‌ی حجم و روش برچسب‌زنی آن با مجموعه داده‌های استاندارد موجود در زبان انگلیسی می‌توان به این نتیجه رسید که مجموعه داده برای انجام پژوهش‌های بیشتر در زمینه‌ی تشخیص موجودیت‌های اسمی مناسب است. در این جا سیستمی اولیه با استفاده از ویژگی‌های ابتدایی برای تشخیص موجودیت‌های اسمی پیشنهاد داده و آن را مورد ارزیابی قرار دادیم. در ادامه‌ی این پژوهش می‌توان تاثیر استفاده از ویژگی‌های پیچیده‌تر مانند نمایش طیفی کلمات و خوشه‌بندی براون را ارزیابی کرد. همچنین می‌توان شبکه‌های یادگیری عمیق پیچیده‌تری را مورد آزمون قرار داد. به ویژه می‌توان مانند پژوهش‌های انجام‌گرفته در زبان انگلیسی از شبکه‌های CNN برای آموزش نمایش طیفی کاراکترها استفاده کرد.

# 7 مراجع

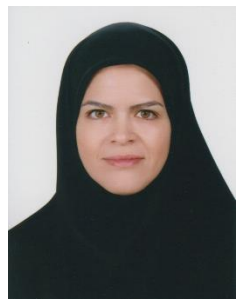

**آزاده شاکری** استادیار دانشکده مهندسی برق و کامپیوتر پردیس دانشکده های فنی دانشگاه تهران است. او مدرک دکترای خود را در سال 1387 از دانشگاه ایلینویز اوربانا - شمپین در آمریکا دریافت کرد. زمینه های پژوهشی مورد علاقه وی مدیریت اطلاعات متنی، بازیابی اطلاعات، متن کاوی، و داده کاوی است. آدرس پست الکترونیکی ایشان عبارت است از: shakery@ut.ac.ir

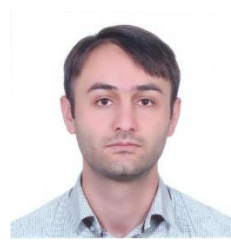

**مهدی محسنی** دانشجوی مقطع دکتری در دانشکده مهندسی برق و کامپیوتر پردیس دانشکده های فنی دانشگاه تهران است. زمینه‌های مورد علاقه وی پردازش زبان و متن‌کاوی است. وی همچنین در حوزه‌های پژوهشی خط و زبان فارسی و تولید منابع زبانی فعالیت دارد. آدرس پست الکترونیکی ایشان عبارت است از: mahdi.mohseni@ut.ac.ir

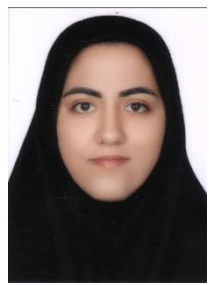

**مهساسادات شهشهانی** دانشجوی مقطع کارشناسی ارشد رشته‌ی مهندسی کامپیوتر-نرم‌افزار در دانشکده مهندسی برق و کامپیوتر پردیس دانشکده‌های فنی دانشگاه تهران است. زمینه‌های مورد علاقه وی شامل بازیابی اطلاعات، پردازش زبان‌های طبیعی، متن‌کاوی و داده‌کاوی است. آدرس پست الکتریکی ایشان عبارت است از: ms.shahshahani@ut.ac.ir